\documentclass{article}
\usepackage{arxiv}
\usepackage[utf8]{inputenc} 
\usepackage[T1]{fontenc}    
\usepackage{hyperref}       
\usepackage{url}            
\usepackage{amsfonts}       
\usepackage{nicefrac}       
\usepackage{lipsum}
\usepackage[all]{hypcap}
\usepackage[noadjust]{cite}
\usepackage[numbered]{bookmark}
\usepackage{silence,etoolbox,graphicx,float,dsfont,color}
\usepackage{regexpatch,microtype,pgffor,bm}
\usepackage{amssymb,mathtools,commath,amsmath,algorithm,algpseudocode}
\usepackage{array,booktabs,subfigure}
\usepackage[capitalize]{cleveref}
\def\BibTeX{{\rm B\kern-.05em{\sc i\kern-.025em b}\kern-.08em
    T\kern-.1667em\lower.7ex\hbox{E}\kern-.125emX}}

\BeforeBeginEnvironment{tabular}{\small\centering}
\graphicspath{{figures/}}

\begin{document}

\title{Enhancing Printed Circuit Board Defect Detection through Ensemble Learning}

\author{ {Ka Nam Canaan Law} \\
	Department of Applied Data Science\\
	San José State University\\
	San José, USA \\
	\And
    {Mingshuo Yu} \\
	Department of Applied Data Science\\
	San José State University\\
	San José, USA \\
	\And
    {Lianglei Zhang} \\
	Department of Applied Data Science\\
	San José State University\\
	San José, USA \\
	\And
    {Yiyi Zhang} \\
	Department of Applied Data Science\\
	San José State University\\
	San José, USA \\
	\And
	{Peng Xu} \\
	Dept. of ISE\\
	Virginia Tech\\
	Blacksburg, USA \\
	\texttt{xupeng@vt.edu} \\
	\And
	{Jun Liu} \\
	Dept. of Computer Engineering\\
	San José State University\\
	San José, USA \\
	\texttt{junliu@sjsu.edu} \\
	\And
	{Jerry Gao} \\
	Dept. of Computer Engineering\\
	San José State University\\
	San José, USA \\
	\texttt{jerry.gao@sjsu.edu} \\
}

\maketitle

\begin{abstract}
The quality control of printed circuit boards (PCBs) is paramount in advancing electronic device technology. While numerous machine learning methodologies have been utilized to augment defect detection efficiency and accuracy, previous studies have predominantly focused on optimizing individual models for specific defect types, often overlooking the potential synergies between different approaches. This paper introduces a comprehensive inspection framework leveraging an ensemble learning strategy to address this gap. Initially, we utilize four distinct PCB defect detection models utilizing state-of-the-art methods: EfficientDet, MobileNet SSDv2, Faster RCNN, and YOLOv5. Each method is capable of identifying PCB defects independently. Subsequently, we integrate these models into an ensemble learning framework to enhance detection performance. A comparative analysis reveals that our ensemble learning framework significantly outperforms individual methods, achieving a $95\%$ accuracy in detecting diverse PCB defects. These findings underscore the efficacy of our proposed ensemble learning framework in enhancing PCB quality control processes.
\end{abstract}

\section{Introduction}

A printed circuit board (PCB) is a medium used to mechanically support and electrically connect various electronic components in a circuit. PCBs play a crucial role in almost every electronic device, from smartphones, computers, and self-driving cars to data centers~\cite{zhang2021multi}. As electronic devices become increasingly complex, and PCBs must be manufactured with higher quality to meet customer demand~\cite{ling2023printed}. Images of PCBs exist in two types: component PCB and bare PCB images. As component PCB images are composed of diverse individual semiconductor cells, it is challenging to use computer vision to detect defects. Bare PCBs do not include through-holes or electronic components, making it easy to identify and recognize defects or missing components~\cite{anoop2015review}. This paper focuses on detecting the defects of bare PCBs. If defects cannot be detected precisely, many produced boards will likely be scrapped eventually, which is both wasteful and costly~\cite{xu2022modeling}. Thus, defect detection is a critical step in quality control to improve yield and profit in manufacturing PCBs.

In current practice, PCB defect detection relies on manual inspection, which is completed by vision tests of experienced workers. As the electronic components become smaller, conventional manual inspection is being replaced by automatic inspection models. Recently, integrating image processing with machine learning algorithms has been a promising direction to build automatic inspection models. However, there are still three challenges during this integration process. First, PCB image datasets are not always clear enough to be used to train models. For example, the surface of PCBs is often covered with a layer of dust or has some black marks of burning, which affects the quality of PCB images. Second, most automatic inspection models rely on only one machine learning algorithm~\cite{ling2023printed}, which makes the inspection results highly dependent on the algorithm training process. Third, building inspection models requires extensive insight and experience, which may not be readily accessible to new users. Thus, a user-friendly inspection model that does not require much prior model knowledge is still lacking.

\begin{figure}[!tbp]
	\centering
	\includegraphics[width=3.5in]{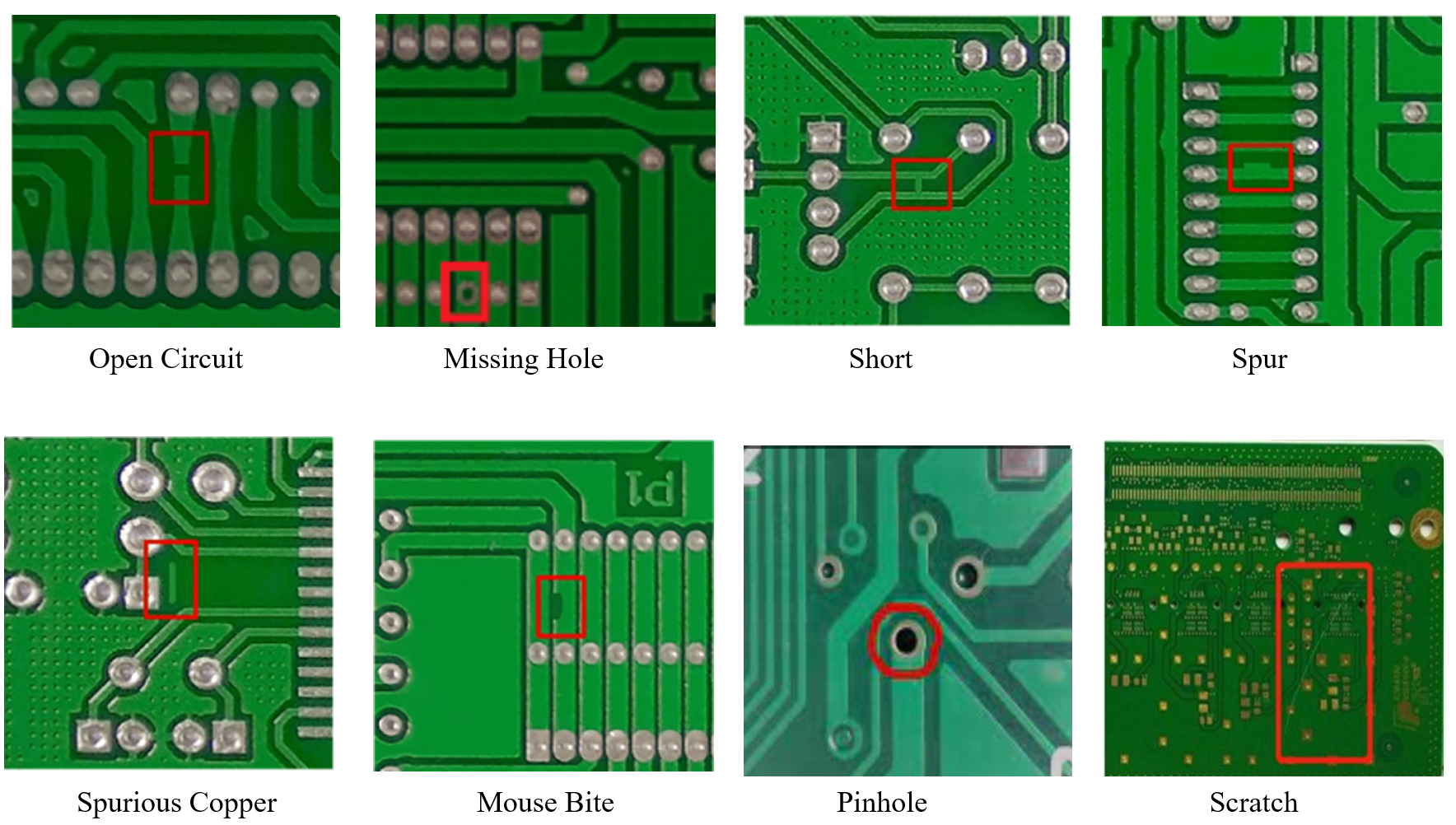}
	\caption{Different types of defects identified on a PCB.}
	\label{fig: samples}
\end{figure}

\begin{figure}[!tbp]
	\centering
	\includegraphics[width=2.2in]{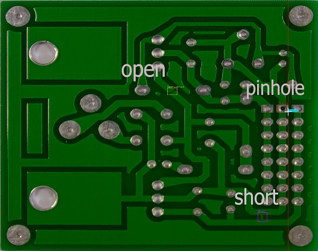}
	\caption{An example of multiple defects displayed on a single PCB.}
	\label{fig: multiple defects}
\end{figure}

The research presented herein seeks to utilize image processing techniques with cutting-edge methods for detecting PCB defects to establish an inspection framework to enhance detection performance. The research scope is to detect eight common defects of PCBs, including \emph{open circuits, short, pinholes, missing holes, scratches, spurious copper, mouse bites, and spurs}~\cite{malge2014pcb}, as shown in \figref{fig: samples}. Also, in \figref{fig: multiple defects}, we demonstrate an example of multiple defects in one PCB. The framework is designed with three components. The first component is data preprocessing. The preprocessed images are used as inputs for training models. The second one is to design an ensemble learning model. We leveraged four different models, including EfficientDet, Faster RCNN, MobileNet SSDv2, and YOLOv5, as base models. These four algorithms are ensembled with a hybrid voting strategy to improve defect detection accuracy. The third one is to develop a web-based interactive application as a pipeline service to users. This application is expected to help users improve the efficiency and quality of defect detection in the PCB manufacturing process.

The remainder of this paper is structured as follows. Section II provides a literature review of PCB defect detection works. Section III describes the details of the proposed inspection framework. Section IV presents our experiments and performance analyses, and finally, the conclusions and future directions are summarized in Section V.

\section{Related Work}
PCB defect detection is an organized process that includes multiple stages designed to pinpoint faults within the PCBs. This process encompasses a range of inspection techniques, data analysis strategies, and quality assurance practices to guarantee the reliability and quality of electronic devices. In this section, we will explore the related work that applies various methods for detecting PCB defects under diverse conditions. 

Automatic optical inspection (AOI) refers to using automated optical systems and technologies to identify faults. A common approach is to visually scan the surface of PCBs for various surface feature defects. These systems employ a combination of hardware and software components to analyze PCBs quickly and accurately and reduce the reliance on manual inspection. The main hardware components of automatic inspection systems are the material and component handling system, illumination system, image acquisition system, and processor~\cite{moganti1995automatic}. It should be noted that each manufacturer of automatic inspection systems utilizes different inspection algorithms and lighting techniques, and each of these systems may have varying strengths and weaknesses depending upon the item/product it inspects ~\cite{taha2014automatic}.

Image processing methods play a critical role in analyzing the captured images of AOI systems. Image processing methods are used to enhance the quality of images, extract relevant features, and identify anomalies that may indicate defects~\cite{ren2015circuit}. For example, dust and strain may reduce image quality during inspection. In general, there are three common methods to deal with these quality issues. The first method is called histogram equalization, which enhances images by remapping the grayscale levels to remove the noise of the data~\cite{anoop2015review}. While this method can make images more detailed with higher quality, it may compress some grayscales during the remapping process. The second method leverages wavelet transform that is widely used in signal and image processing~\cite{zhu2018printed,jing2024computer,Tirumala2024building}. This method has characteristics such as low entropy, multi-resolution, and decorrelation. It is also good at removing image noises because it can overcome the issue of window size not varying with frequency. However, the selection of wavelet basis is complex, and the results of wavelet basis analysis are different.

The third method is based on machine learning methods, such as convolutional neural networks (CNN), generative adversarial networks (GAN), and random forest regression (RFR)~\cite{xu2023ucb}. For example, Gampala et al.~\cite{gampala2020deep} used a fully connected neural network (FCNN) to solve motion blur problems and enhance the quality of images. The input can be partial images that save the storage. However, there is some lack of noise training for multi-frame deblurring. Lin and Menendez~\cite{lin2019image} proposed an image-generating and processing method that denoises the target region of images. This method can be applied in different pad positions. However, some defects in generated images remain, such as blurred image edges and low resolution. Mironic~\cite{mironicua2020generative} presented a GAN-based method that can remove dust and scratches from the scanning of films. This method achieved better performance than previous solutions in terms of PSNR and SSIM quality metrics. However, GAN can result in a blurry image after dirt cleaning, and it is possible to miss some scratches.

Building a deep learning model typically involves several stages, including data collection, framework design, model training, and performance evaluation~\cite{deng2014deep, xu2020reinforcement, xu2016feedrate, rangwala2021deeppastl, liu2022intermittent}. CNN-based models have achieved a lot of success in computer vision fields, such as image classification, recognizing faces, instance segmentation, and tracking targets~\cite{saberironaghi2023defect}. The success of CNN-based models can be attributed to two advantages~\cite{ling2023printed}. First, CNN-based methods are able to extract image features efficiently, simplifying the image pre-processing process so that the detection accuracy and speed can be promoted effectively. Second, the methods are robust to the environment and noise.

Recently, many architectures have been applied in a wide range of industrial defect detection, such as PCB defect detection~\cite{ling2023printed}, cosmetic defect detection~\cite{zhang2021cs}, and automotive defect defection~\cite{du2019approaches}. Various models have been proposed to solve the detection problems, including ResNet, RCNN, AlexNet, CNN-VGG16, and RB-CNN. For example, Hu and Wang~\cite{hu2020detection} proposed Faster RCNN with ResNet50 to detect six types of PCB defects: solder ball, pinhole, spur, mouse bite, and short. Their model has better accuracy than other benchmark methods but is sensitive to different datasets. Lu et al.~\cite{lu2020fics} presented AlexNet to recognize defects in PCB images. A ReLU Nonlinearity function is used to solve the gradient vanishing problem in deep networks. However, this model has higher requirements on GPU. Ge et al.~\cite{ge2022yopcb} proposed an improved model YOPCB for PCB images. However, an annotation step is required before model training. Revaud et al. developed a model called~\cite{revaud2019learning} CNN-VGG-16 to detect five types of defects: short, mouse bite, coppers, open, and spur. However, their model is limited by the feature selection algorithm.

All models above have their advantages and disadvantages, and their performance also depends on specific datasets. For example, Faster-RCNN with ResNet50-FPN outperforms RetinaNet and YOLOv3 in some datasets~\cite{hu2020detection}. Zhang et al.~\cite{zhang2021cs} showed that CS-ResNet has the highest sensitivity, G-mean, and the lowest misclassification cost. Zhang et al.~\cite{zhang2018improved} demonstrated that VGG obtains the highest mAP and high precision and recall, which means it has high accuracy. Therefore, how to reduce the impact of errors caused by datasets or training processes is critical in practice. Ensemble learning is an effective technique that has increasingly been adopted to combine multiple learning models to improve overall prediction accuracy~\cite{dietterich2000ensemble}. These ensemble techniques have the advantage of alleviating the small sample size problem by incorporating multiple learning models to reduce the potential for overfitting the training data~\cite{yang2010review, xu2020expert}. For example, voting is a common technique that determines the final output by the majority of all models' results ~\cite{CasadoGarcia19}. As an attempt, Li et al.~\cite{li2020automatic} proposed a deep ensemble model combining hybrid YOLOv2 and Faster R-CNN to achieve PCB defect detection. However, they concentrate solely on one type of defect. Consequently, there remains a need for a comprehensive ensemble model that can address multiple types of PCB defects.

\section{Ensemble Learning for PCB Defects Detection and Classification}
\label{Section III}

In this section, we present the proposed framework, consisting of data preprocessing methods and an ensemble learning model for defect detection. Specifically, this work focuses on detecting the following types of defects: \emph{mouse bite, pinhole, spurious copper, open circuit, spur, scratch, short, missing hole}. 

Data preprocessing is a crucial step in building models for defect detection. It involves preparing and transforming raw data into a format suitable for the training requirements of methods. We use PCB images from electronics and semiconductor industries as primary datasets to collect raw defect information. During the data preprocessing process, there are three challenges in dealing with PCB images. The first challenge comes from the diversity of data sources. All PCB images are collected from outside sources, and each outside source has its format standards, such as image size and resolution. The second one is related to image quality. The raw images are collected from a camera or computer vision and may contain noise, artifacts, or irrelevant information. Such noise information could interfere with the model's ability to detect defects. For example, if an image is blurred or its details are not precise, it would be hard to distinguish an open circuit from a mouse bite according to this image. The third one is about the ground truth information of PCB defects. After raw images are collected, the ground truth information of PCB defects is unknown in each image. While some PCBs have no defects, others may have multiple defects. How to capture defect information from images is critical to the performance evaluation. To handle these challenges, we leverage different data preprocessing methods to transform collected datasets.

\begin{figure}[!tbp]
	\centering
	\includegraphics[width=3.5in]{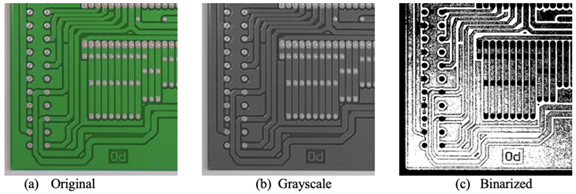}
	\caption{An example illustrating the image binarization process.}
	\label{fig: 1}
\end{figure}

To solve the first challenge, we combine and uniform all datasets from different data sources. Because the high resolution of image data affects the detection accuracy of the model and the cost of computing resources, all image sizes are uniformed to $600 \times 600$ pixels in our framework. The denoising method is also used to enhance object edges and improve the resolution of images. We choose the open Python source package Scikit-image to make image sizes uniform. In order to avoid contrast disturbance and unify all input data formats, we binarize all color images into black and white. First, we convert all color images into grayscale.
Lastly, all images are unified to the same color scale and format before labeling, as shown in \figref{fig: 1}.

\begin{figure}[!tbp]
	\centering
	\includegraphics[width=3.5in]{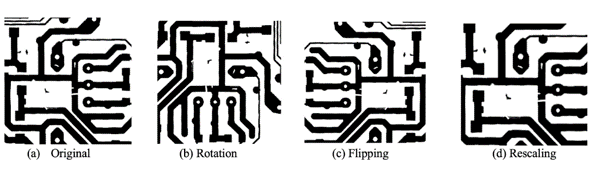}
	\caption{Samples of Data Augmentation for Binarized Image.}
	\label{fig: 8}
\end{figure}

To solve the second challenge, we leverage data augmentation to enhance image quality by applying various transformations to the original images. The primary goal of data augmentation is to introduce variations in the training data that mimic real-world scenarios. By doing so, data augmentation helps improve the model's ability to generalize to unseen data and reduces the risk of overfitting. Some common image data augmentation techniques include random rotating, rescaling, zooming, color modification, image flipping, etc. These methods are common used in different applications, such as image classification, visual detection, and object segmentation~\cite{wang2017effectiveness}. The methods used in our approach include random rotation, image flipping, brightness changing, and image rescaling. All these techniques are realized with the package Scikit-image. Other techniques are reserved for future work. We utilize flipping and rotation to create diverse backgrounds. All images are flipped or rotated at different angles to simulate variations in object orientation. We use brightness changing to modify the original images' brightness to simulate different luminance situations. Image rescaling is leveraged to increase the various scopes of the same defects. \figref{fig: 8} shows sample images after different data argumentation operations.

As to the third challenge, we label each image with its corresponding defects to annotate the ground truth information by utilizing the tool OpenLabeling Annotator~\cite{8594067}. The location of the defects and the types of defects are bounded by the rectangle boxes and displayed on the top box.

\begin{figure*}[!tbp]
	\centering
	\includegraphics[width=6in]{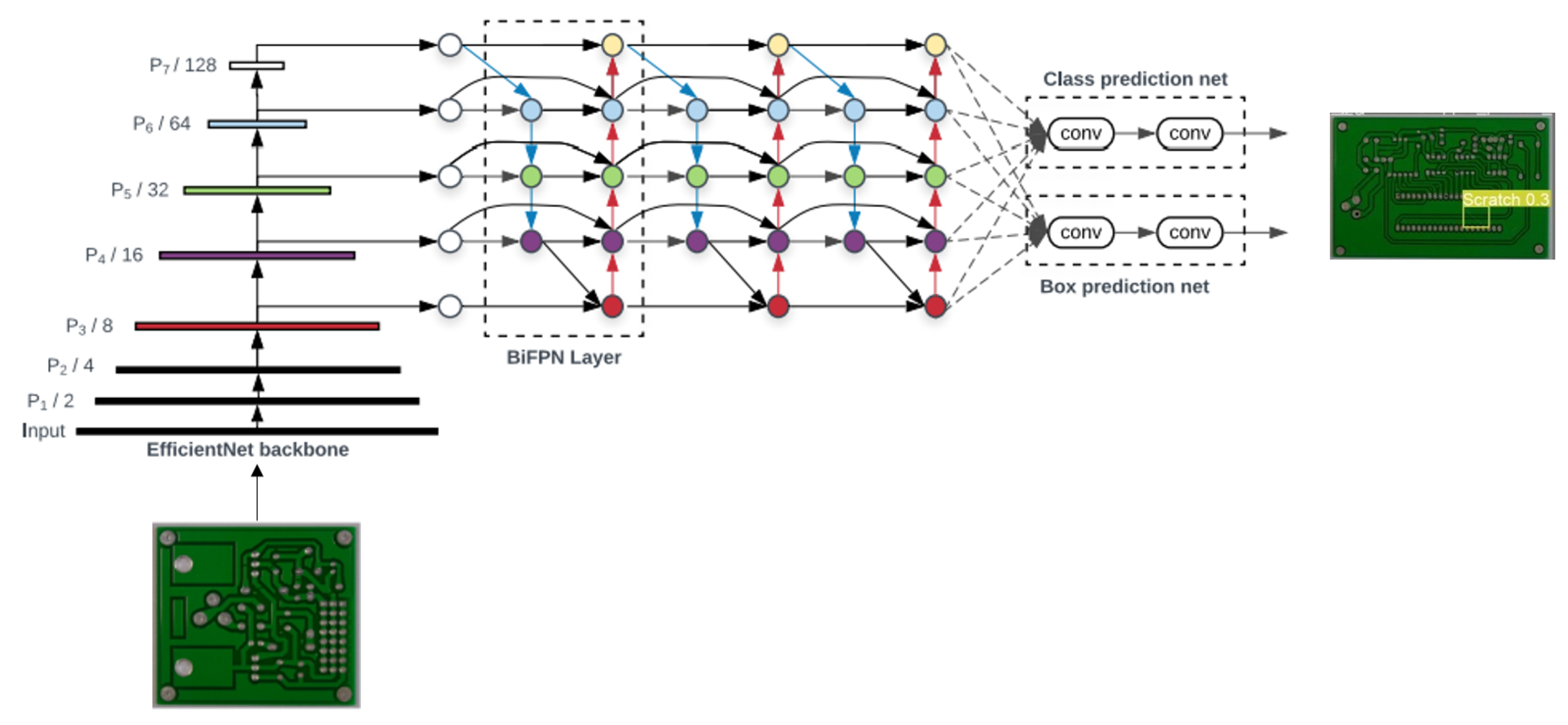}
	\caption{The overview of EfficientDet-based PCB defects detection architecture. The architecture contains three components, including backbone, feature fusion, and class/box network. An ImageNet-pretrained EfficientNets is employed as the backbone network.}
	\label{fig: 14}
\end{figure*}

\begin{figure*}[!tbp]
	\centering
	\includegraphics[width=5in]{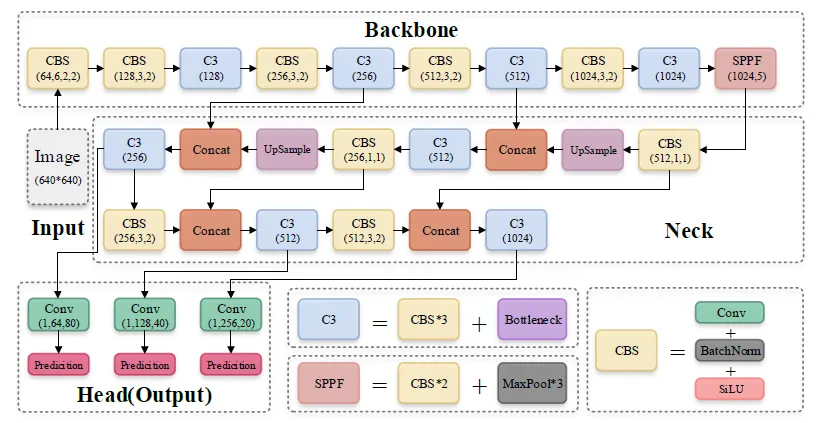}
	\caption{The overview of YOLOv5 network architecture used for PCB defects detection. YOLOv5 leverages a single-stage detector architecture, incorporating spatial pyramid pooling and Cross Stage Partial networks to enhance both speed and accuracy. These architectural enhancements are particularly effective for our PCB defects detection problem, which requires high precision and efficiency, where rapid and accurate flaw identification is critical.}
	\label{fig: YOLOv5}
\end{figure*}

Then, an ensemble learning model is presented to address the problem of robust PCB defect detection in this sub-section. This model is implemented in two phases. The first phase is to train each base model independently and perform model optimization using cross-validation. Each trained base model is used to make predictions about PCB defects. We select four deep learning models as base models, including EfficientDet, Faster RCNN, MobileNet SSDv2, and YOLOv5. All those models will be discussed in Section~\ref{Fundamental Models}. Second, all prediction results are ensembled using a hybrid voting strategy, as presented in Section~\ref{Ensemble Learning}.

\subsection{Fundamental Models for Ensemble Learning}
\label{Fundamental Models}

\paragraph{EfficientDet}
EfficientDet is a one-stage object detection model developed by Tan et al.~\cite{8594067}. The architecture of EfficientDet contains three components, including backbone, feature fusion, and class/box network. An ImageNet-pretrained EfficientNets is employed as the backbone network. The bi-directional feature pyramid network (BiFPN) proposed in~\cite{tan2020efficientdet} serves as the feature network, which takes level 3-7 features from the backbone network and repeatedly applies top-down and bottom-up bidirectional feature fusion. These fused features are fed to a class and box network to produce object class and bounding box predictions respectively. With its ability to scale up model complexity while maintaining memory and computational constraints, EfficientDet can achieve an optimal balance between accuracy, efficiency, and computational cost. This advantage is attributed to weighted feature fusion and compound scaling methods. As EfficientDet achieves the best performance on some datasets, such as COCO and Pascal VOC, over other models, it is considered in our ensemble learning model. The original work of EfficientDet provides seven variants from EfficientDet-D0 to EfficientDet-D7. Even though EfficientDet-D7 comes with higher computational requirements and resource constraints, it has the highest accuracy and better generalization performance among all variants. Therefore, we adopt this variant in our model.

\paragraph{MobileNet SSDv2}
MobileNet-SSDv2 is a lightweight object detection model that combines the backbone network MobileNet with the one-stage object detector Single Shot Multibox Detector (SSD) for real-time object detection tasks~\cite{chiu2020mobilenet}, as shown in \figref{fig: 14}. MobileNet utilizes depthwise separable convolutions to reduce the number of parameters and computational costs while maintaining strong feature representation capabilities~\cite{howard2017mobilenets}. This allows MobileNet-SSDv2 to achieve real-time inference speed without compromising on detection accuracy. SSD is a single deep neural network that discretizes input images into a set of default boxes at multiple scales and predicts scores for each object category in each default box~\cite{liu2016ssd}. This multi-scale approach enables Single Shot Multibox Detector to detect objects of various sizes and aspect ratios in a single forward pass, making it well-suited for real-time applications. By optimizing MobileNet and Single Shot Multibox Detector, MobileNet-SSDv2 achieves better performance in object detection while maintaining low computational overhead. With its compact architecture and real-time inference speed, MobileNet-SSDv2 is widely adopted for various image recognition applications, especially for resource-constrained devices. .

\paragraph{Faster RCNN}
Faster R-CNN (Region-based Convolutional Neural Network) is a two-stage object detection model introduced by Shaoqing Ren et al. in 2016~\cite{ren2015faster} as an improvement over earlier object detection models R-CNN and Fast R-CNN. Faster RCNN consists of three key components: region proposal network (RPN), region of interest (ROI) pooling, and object detection network. RPN is the most critical part of Faster R-CNN as it can share full-image convolutional features with the detection network to enable nearly cost-free region proposals. As RPNs are trained to generate high-quality region proposals, Faster RCNN performs better than those earlier models. RPN also shares the same last convolutional layer with the Fast RCNN when processing the output image feature map. So, the data training is finished only once, which saves training time. Due to the introduction of ROI Pooling, Faster RCNN can detect images of arbitrary sizes. So, it’s more flexible when applying object detection to the images. Thus, it is also utilized in our ensemble learning model.

\paragraph{YOLOv5}
The framework of YOLO5 has three components, including backbone, neck, and prediction networks~\cite{liu2022sf}, as shown in Figure ~\ref{fig: YOLOv5}. CSPDarknet53 is used to extract features in the backbone network to reduce memory and computational usage. Three resolution feature maps were extracted as the output of the backbone network. Feature pyramid network (FPN) and pyramid attention network (PAN) are utilized in the neck network. The basic idea of FPN is to up-sampling the output feature maps to generate multiple new feature maps for detecting different scale targets. The prediction networks use anchor boxes to predict the detection labels based on the feature maps. This model also has some other functional components, i.e., auto-anchoring and default augmentation, to achieve better detection performance.
It has also gained higher performance than its previous version in certain situations. This model also acts as a suitable object detector for small objects. Therefore, YOLOv5 is selected as part of our ensemble learning model.

\begin{figure}[!tbp]
	\centering
	\includegraphics[width=2.8in]{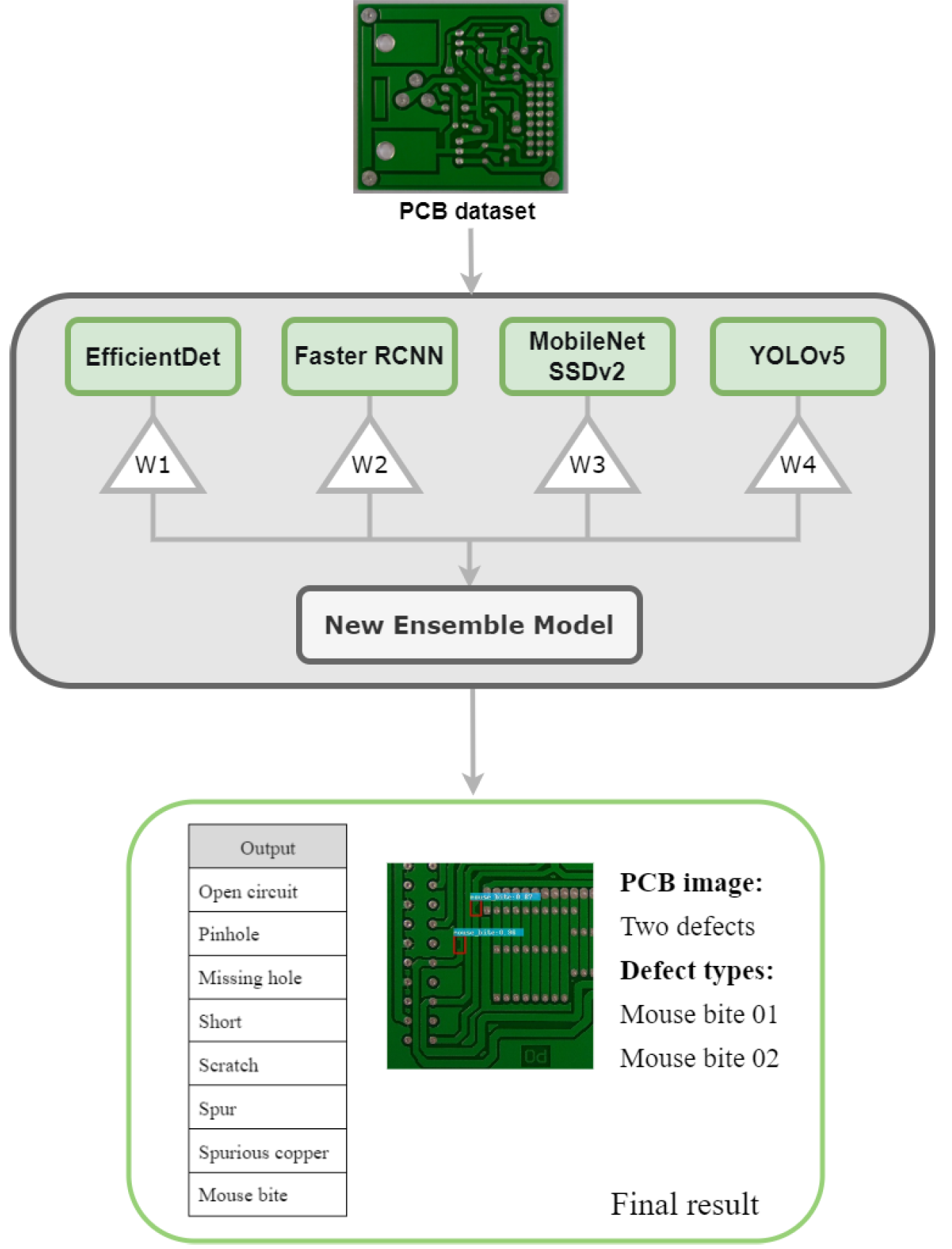}
	\caption{The proposed ensemble learning architecture by utilizing four state-of-the-art object detection algorithms: EfficientDet, MobileNet SSDv2, Faster R-CNN, and YOLOv5.}
	\label{fig: ensemble}
\end{figure}

\subsection{Ensemble Learning Model}
\label{Ensemble Learning}

In this work, we utilize a hybrid voting strategy that ensembles the previous four models in two steps. First, all object boxes identified by the base models are considered valid. Thus, the results of all base models are saved in the ensemble learning model, resulting in no information loss. Second, for each object box, the type of a PCB defect is ensembled with a consensus voting strategy~\cite{CasadoGarcia19}. Recognizing the unique strengths and performance characteristics of each model, we can assign distinct weights to each base model, denoted as \( w_1, w_2, w_3, \) and \( w_4 \). These weights can be determined based on the performance accuracy and reliability of each base model during the validation phase. The consensus score \( s \) for detecting an object class across all models is computed as follows:
\begin{equation}
\begin{split}
s = & w_1 \cdot p_{\text{EfficientDet}} + w_2 \cdot p_{\text{MobileNet}} \\
    & + w_3 \cdot p_{\text{FasterRCNN}} + w_4 \cdot p_{\text{YOLOv5}}
\end{split}
\end{equation}
where \( p_{\text{EfficientDet}} \), \( p_{\text{MobileNet}} \), \( p_{\text{FasterRCNN}} \), and \( p_{\text{YOLOv5}} \) are the confidence scores from each respective model for a particular object class. The final ensemble decision to accept a detection as valid is based on the consensus score \( s \). This weighted approach allows us to capitalize on the strengths of each model, thereby enhancing the overall detection accuracy. The overview of the framework is shown in \figref{fig: ensemble}.

\begin{table*}[!tbp]
	\centering
	\caption{Comparative Performance Analysis of the Ensemble Model and the Four Individual Base Models.}
	\label{tab: comparison}
	\begin{tabular}{cccccc}
		\toprule
		Model  & $\text{mAP}_{0.5:0.95}$ &  $\text{mAP}_{0.5}$ & Accuracy & Training Time (hour) & Average Run Time (second)\\
		\midrule
		FastRCNN       & 30.0\%  & 65.0\%             & 93\%  & 5    & 1.66    \\
		MobileNet      & 29.8\%  & 67.3\%             & 88\%  & 4.5  & 1.43     \\
		EfficientDet   & 29.3\%  & 55.8\%             & 83\%  & 5    & 0.36     \\
		YOLOv5         & 34.5\%  & 79.6\%             & 94\%  & 3.7  & 0.06     \\
		\emph{Ensemble Model} & \textbf{35.3\%}  & \textbf{80.3\%}             & \textbf{95\%}  & N/A  & 3.91     \\
		\bottomrule
	\end{tabular}
\end{table*}

\section{Experimental Results}

In this section, we will present the details of the experimental design to verify the detection performance of our ensemble learning model.

Three data sources are collected to support model building and testing in this experiment. The first data source is from~\cite{huang2020hripcb}. This dataset contains $692$ images that cover six defect types: a missing hole, mouse bite, open circuit, short, spur, and spurious copper. All the image samples are based on ten template PCBs checked manually by industry experts to be used as standards. The second data source is from~\cite{ding2019tdd}, and the images are cropped to $600 \times 600$. This dataset has $9,920$ images for training data and $2,508$ images for testing data. The third data source is from~\cite{tang2019online}. There are $1,500$ image pairs in size $640 \times 640$, and each pair consists of defect-free and defective versions. This dataset includes the same six defects as the first data source. As two types of defects, including \emph{pinhole} and \emph{scratch}, are missing, we manually add these two types to the original PCB images. Thus, all eight types of PCB defects are covered in our datasets. All these datasets are processed with the data preprocessing methods to obtain a complete dataset with a unified image format in Section~\ref{Section III}.

Next, the preprocessed dataset is split into training, testing, and validation sets. While $70\%$ of the dataset is allocated as the training set, testing and validation sets constitute $15\%$ of the dataset separately. To improve the training performance, the splitting process is enhanced by a balanced allocation method. That is, even though different types of defects have different defect counts, all types of defects are adjusted to follow the $70\% - 15\% - 15\%$ rule.

With the collected sets, we utilized the four different models independently as benchmark models of PCB defect detection. Each model is trained with the same image dataset of PCBs and then detects and classifies the problematic area. Two model training techniques are applied to enhance model performance. The first one is to tune hyper-parameters, including learning rate, batch size, and optimizer. For EfficientDet, the batch size is set as $16$, and the learning rate is $0.001$. For MobileNet, the batch size is set as $12$. For Faster RCNN, the batch size is set as $8$, and the learning rate is $0.01$. For YOLOv5, the batch size is $16$, and the learning rate is $0.01$ with a decay method. An RMS prop optimizer is chosen for MobileNet, and momentum optimizers are selected for three other models to optimize network parameters. The second one is to validate detection results by visualizing defect types and confidence values on images. Then, we utilized the proposed ensemble learning model to improve the detection performance with regard to benchmark models. We use mAP (mean Average Precision)~\cite{revaud2019learning} as the metric to assess the detection performance of different models. It calculates the average precision while stating the accuracy of the PCB defect detection as follows:

\begin{equation}
	\text{mAP} = \frac{\sum_{i=1}^C \text{AP}_i}{C}
\end{equation}
where $C$ is the total number of samples and $\text{AP}_i$ is the Average Precision of $i$th sample. As the precision depends on the Intersection over the Union (IoU) threshold, we leverage two mAP metrics. The first metric $\text{mAP}_{0.5}$ is the mAP under the IoU threshold $0.5$. The second metric $\text{mAP}_{0.5:0.95}$ is the mAP under multiple IoU thresholds, ranging from $0.50$ to $0.95$, with a step size of $0.05$.
Other metrics used in the experiment are accuracy, precision, and recall, which are defined as follows:
\begin{equation}
    \text{accuracy} = \frac{\texttt{TP + TN}}{\texttt{TP + TN + FP + FN}}.
\end{equation}

\begin{equation}
    \text{precision} = \frac{\texttt{TP}}{\texttt{TP + FP}}.
\end{equation}

\begin{equation}
    \text{recall} = \frac{\texttt{TP}}{\texttt{TP + FN}}.
\end{equation}
where \texttt{TP} is true positive, \texttt{TN} is true negative, \texttt{FP} is false positive, and \texttt{FN} is false negative.

We use two metrics, including training time and average run time, to evaluate the time performance of defect detection models. The training time only refers to the training time length of a detection model. The average run time means the average time length it takes to process a single image.

\subsection{Experimental Results}

We utilize the four base models sequentially to detect PCB defects. The first model, EfficientDet, is trained with all sample images. The loss value became stable after the 30k round of the training process. The trained model can identify all eight types of PCB defects. Among them, the defect \emph{pinhole} has the most significant F1 score.

The second model, MobileNetV2, is also utilized on the same dataset. It can be observed that the loss of the model decreases from epoch $0$ to epoch $30$ during the whole training process, which indicates that the model works well with defect detection on PCB with coordinates. While the object loss dropped to the lowest value around epoch $10$, it bounced back and increased significantly during epoch $10$ to $30$. This indicates poor prediction performance. More hyper-parameter tuning work may boost this outcome. The classification loss is constantly decreasing, which shows the model is well-trained to predict the types of defects after locating them on PCB.

The third model, Faster RCNN, is trained for $7,000$ steps with three optimization methods, including momentum-based gradient optimizer, Adam optimizer, and RMSProp optimizer. We also evaluate the model with the mAP at $50$ and classification loss. The Adam optimizer has the lowest loss value, but the mAP is extremely low. This issue might be caused by hyperparameters such as batch size or learning rate. However, as with other optimization methods, the mAPs are still increasing at $7,000$ steps, while losses have become stable. The trained model can identify all eight PCB defects.

\begin{figure}[!tbp]
	\centering
	\includegraphics[width=3in]{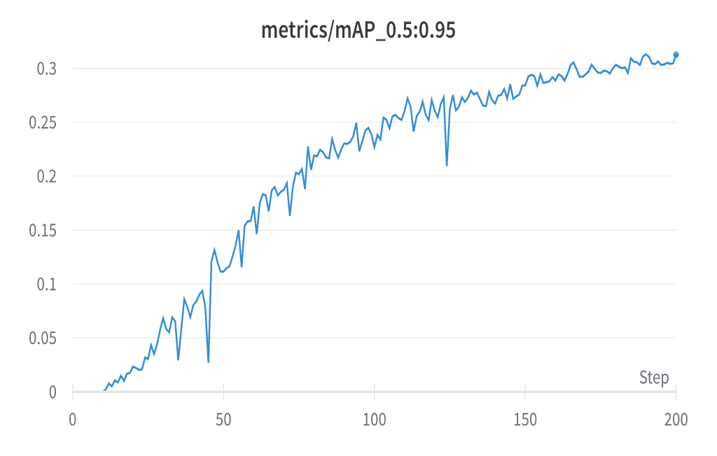}
	\caption{The training performance (metric: mAP) of YOLOv5. $x$ axis is training step, and $y$ axis is mAP.}
	\label{fig: 26}
\end{figure}

The fourth model is YOLOv5. As shown in \figref{fig: 26}, the mAP score increased at a smooth pace during the training epochs except for the interval from $15$ to $18$. The loss value decreases and becomes stable after the $20$th epoch. The precision is constantly increasing and is more significant than $60\%$ at the end of epoch $30$. More training epochs may result in better training performance.

\begin{figure}[!tbp]
	\centering
	\includegraphics[width=3.3in]{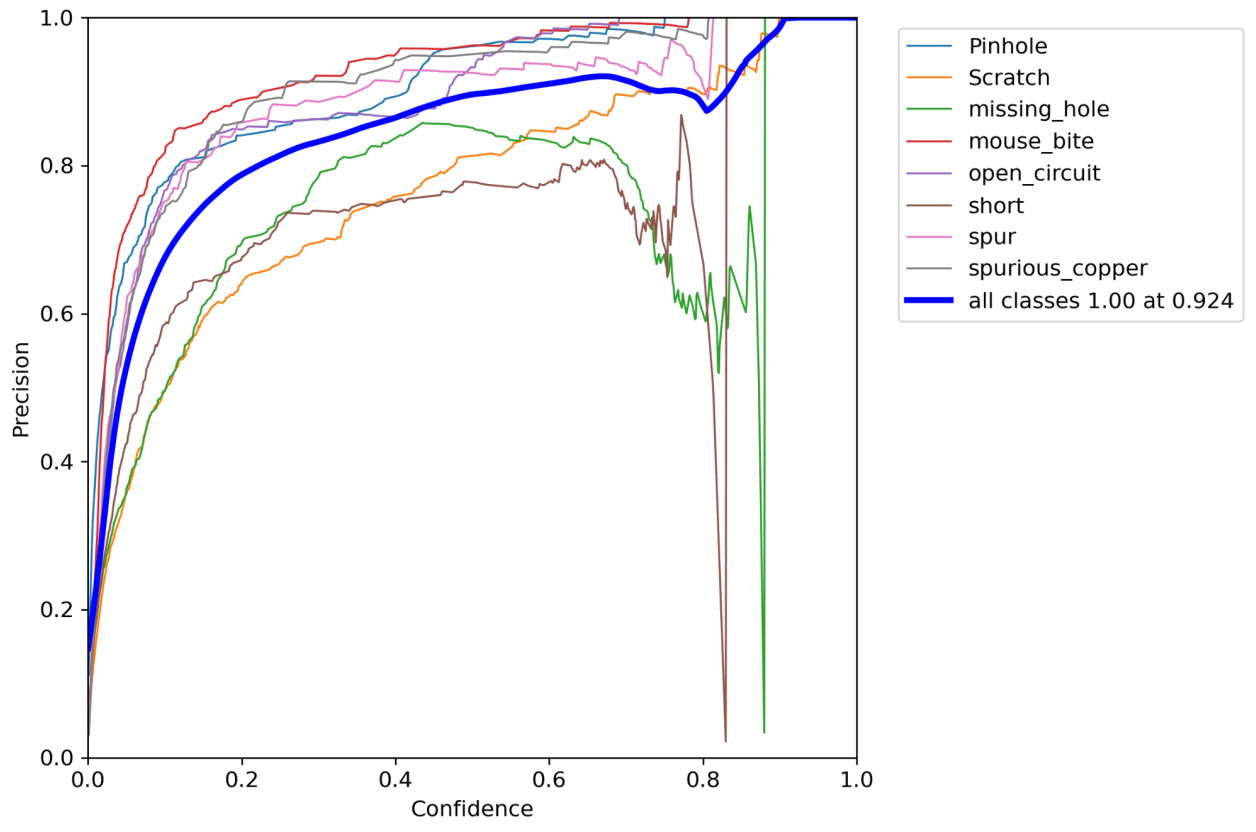}
	\caption{Precision values of eight PCB defects predicted by YOLOv5.}
	\label{fig: 28}
\end{figure}

The performance of all base models is summarized in Table \ref{tab: comparison}. Among all base models, YOLOv5 has the optimal overall performance and the least average run time. But, its prediction performance for the defects \emph{missing hole} and \emph{short} is slightly worse compared with other base models, as shown in Figure ~\ref{fig: 28}. In contrast, the proposed ensemble learning model can aggregate the strengths of all base models for defect detection. The performance of the proposed ensemble learning model is shown in the last row. It can be observed that its mAP values are larger than all base models, and its accuracy also outperforms others. As our ensemble model needs the results of all base models, its average run time is approximately the sum of the average run times of base models, which is expected.

\section{Conclusion and Future Work}

In this paper, we introduced a novel inspection framework designed to detect defects in printed circuit boards (PCBs). This comprehensive approach involves stages of data collection, preprocessing, model development, evaluation, and the development of a web application. We utilized four established models, i.e., EfficientDet, MobileNet, FasterRCNN, and YOLOv5, to identify PCB defects independently. These models were integrated using a hybrid voting strategy in our ensemble learning model, which achieved a detection accuracy of $95\%$. The proposed method is versatile enough to detect various types of PCB defects and can be adapted for defect detection in other manufacturing contexts, demonstrating its potential as a generalized approach to enhance performance. This adaptability underscores its capability to improve defect detection across different manufacturing scenarios.

Future research directions for this framework include extending its capabilities to support real-time video inputs and exploring other ways to improve defect detection performance. These initiatives will help to advance the field of PCB quality control and broaden the applicability of our methods.

\bibliographystyle{IEEEtran}
\bibliography{PCB}

\end{document}